\documentclass{trbunofficial}

\usepackage{graphicx}
\usepackage{booktabs}
\usepackage{svg}
\usepackage{longtable}
\usepackage{amsmath}
\usepackage{amssymb}
\usepackage{threeparttable}
\usepackage{multirow}
\usepackage{bbding}
\RequirePackage{geometry}
\usepackage[hidelinks]{hyperref}

\AuthorHeaders{Huang, Yang and Yuan}
\title{Leveraging Intra-Period and Inter-Period Features for Enhanced  Passenger Flow Prediction of subway stations}

\author{%
  \textbf{Xiannan Huang}\\
  Phd Candidate,\\
  Key Laboratory of Road and Traffic Engineering, Ministry of Education at Tongji University,\\
  4800 Cao’an Road, Shanghai, 201804, China\\
   Emali: huang\_xn@tongji.edu.cn\\
  \hfill\break%
  \textbf{Chao Yang, Corresponding Author}\\
  Professor,\\
  Key Laboratory of Road and Traffic Engineering, Ministry of Education at Tongji University,\\ Urban Mobility Institute, Tongji University,\\
  4800 Cao’an Road, Shanghai, 201804, China\\
  Email: tongjiyc@tongji.edu\\
  \hfill\break
  \textbf{Quan Yuan}\\
  Distinguished Researcher,\\
  Urban Mobility Institute, Tongji University,\\
  1239 Siping Road, Shanghai, 200082, China\\
  Email: quanyuan@tongji.edu.cn
}

\begin{document}
\begin{center}
    \huge \textbf{Leveraging Intra-Period and Inter-Period Features for Enhanced Passenger Flow Prediction of Subway Stations} \\[1em]
    
\end{center}
    \textnormal{Xiannan Huang}\\
    PhD Candidate,\\
    Key Laboratory of Road and Traffic Engineering, Ministry of Education at Tongji University,\\
    4800 Cao’an Road, Shanghai, 201804, China\\
    Email: huang\_xn@tongji.edu.cn\\
    \hfill\break
    \textnormal{Chao Yang, Corresponding Author}\\
    Professor,\\
    Key Laboratory of Road and Traffic Engineering, Ministry of Education at Tongji University,\\
    Urban Mobility Institute, Tongji University,\\
    4800 Cao’an Road, Shanghai, 201804, China\\
    Email: tongjiyc@tongji.edu\\
    \hfill\break
    \textnormal{Quan Yuan}\\
    Distinguished Researcher,\\
    Urban Mobility Institute, Tongji University,\\
    1239 Siping Road, Shanghai, 200082, China\\
    Email: quanyuan@tongji.edu.cn

\vspace{2em} 
\noindent Word Count: 5577 words + 4 table(s) × 250 = 6577 words
\newpage
\section{Abstract}
Accurate short-term passenger flow prediction of subway stations plays a vital role in enabling subway station personnel to proactively address changes in passenger volume. Despite existing literature in this field, there is a lack of research on effectively integrating features from different periods, particularly intra-period and inter-period features, for subway station passenger flow prediction. In this paper, we propose a novel model called \textbf{M}uti \textbf{P}eriod \textbf{S}patial \textbf{T}emporal \textbf{N}etwork \textbf{MPSTN}) that leverages features from different periods by transforming one-dimensional time series data into two-dimensional matrices based on periods. The folded matrices exhibit structural characteristics similar to images, enabling the utilization of image processing techniques, specifically convolutional neural networks (CNNs), to integrate features from different periods. Therefore, our MPSTN model incorporates a CNN module to extract temporal information from different periods and a graph neural network (GNN) module to integrate spatial information from different stations. We compared our approach with various state-of-the-art methods for spatiotemporal data prediction using a publicly available dataset and achieved minimal prediction errors. The code for our model is publicly available in the following repository: https://github.com/xiannanhuang/MPSTN


\newpage

\section{Introduction}
Forecasting passenger flow at subway stations is of significant importance in the operation of subway system. By accurately predicting the passenger flow for future intervals such as 15 minutes, 30 minutes, 45 minutes, or 1 hour, appropriate operation strategies can be implemented to prevent overcrowding and mitigate potential safety hazards.

Predicting the passenger flow of subway stations in a city for a future time period can be regarded as a spatiotemporal prediction problem \cite{Liu2020PhysicalVirtualCM}. Typically, handling such problems involves analyzing two aspects. The first aspect is the temporal dimension, where the passenger flow of a specific station in the future time period is influenced by its past passenger flow over a certain time span. The second aspect is the spatial dimension, where the passenger flow of different stations is interrelated. Consequently, existing methods often consider the spatial relationships using graph neural networks (GNNs) and incorporate the temporal dependencies using recurrent neural networks (RNNs). We acknowledge that these approaches are reasonable, but there is room for improvement.

In our paper, we focus on how to integrate the features of different periodicities in passenger flow of subway stations to improve the accuracy of flow prediction. Here, different periodicities refer to different days, as it is evident that each day exhibits a clear periodic pattern in subway passenger flow data. Therefore, our approach involves utilizing not only the data from the previous time steps within the same day but also the data from several previous days or even weeks to predict the passenger flow for a future time period, such as the next hour.

\begin{figure}[!h]
    \centering
    \includegraphics[width=\textwidth,height=0.26\textwidth]{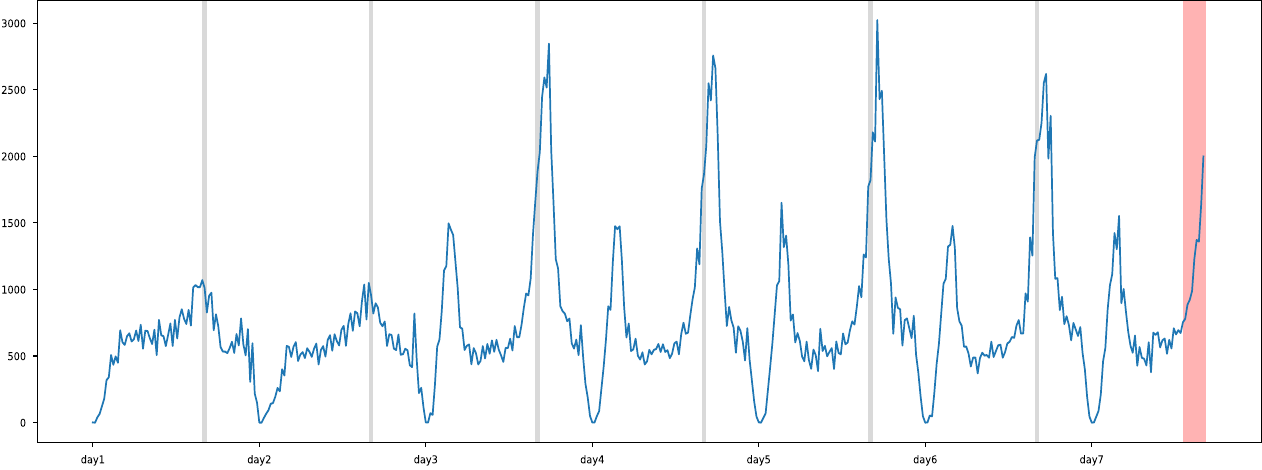}
    \caption{The inflow of a station in the past seven days. The data in the red area represents the inflow of several time intervals immediately preceding the future time to be predicted, while the data in the grey area represents the inflow at the same time interval to be predicted on several previous days.}
    \label{fig1}
\end{figure}

However, it should be noted that using data from several preceding days or weeks to predict traffic flow is an explored topic \cite{Li2022ImprovingSB,Guo2019AttentionBS}. In such cases, the methods typically treat data from different periodicities as separate time series and just as illustrated in Fig.\ref{fig1}. If the prediction of future passenger flow is desired, the data in red area is the passenger flow at several preceding time intervals of the time interval to be predicted and is can be regarded as a time series. While the data enclosed in the grey areas represents passenger flow at same time interval to be predicted on several previous days, and they can be concatenated chronologically as a time series. By treating each day as a distinct period, the data in the red area demonstrates the trend within a period, while the data in the grey areas reflects the trends between periods. By integrating and modeling these two time series data, the information within and between periods can be effectively integrated and more precise predictions can be obtained.
\begin{figure}[!h]
    \centering
    \includegraphics[width=\textwidth,height=0.68\textwidth]{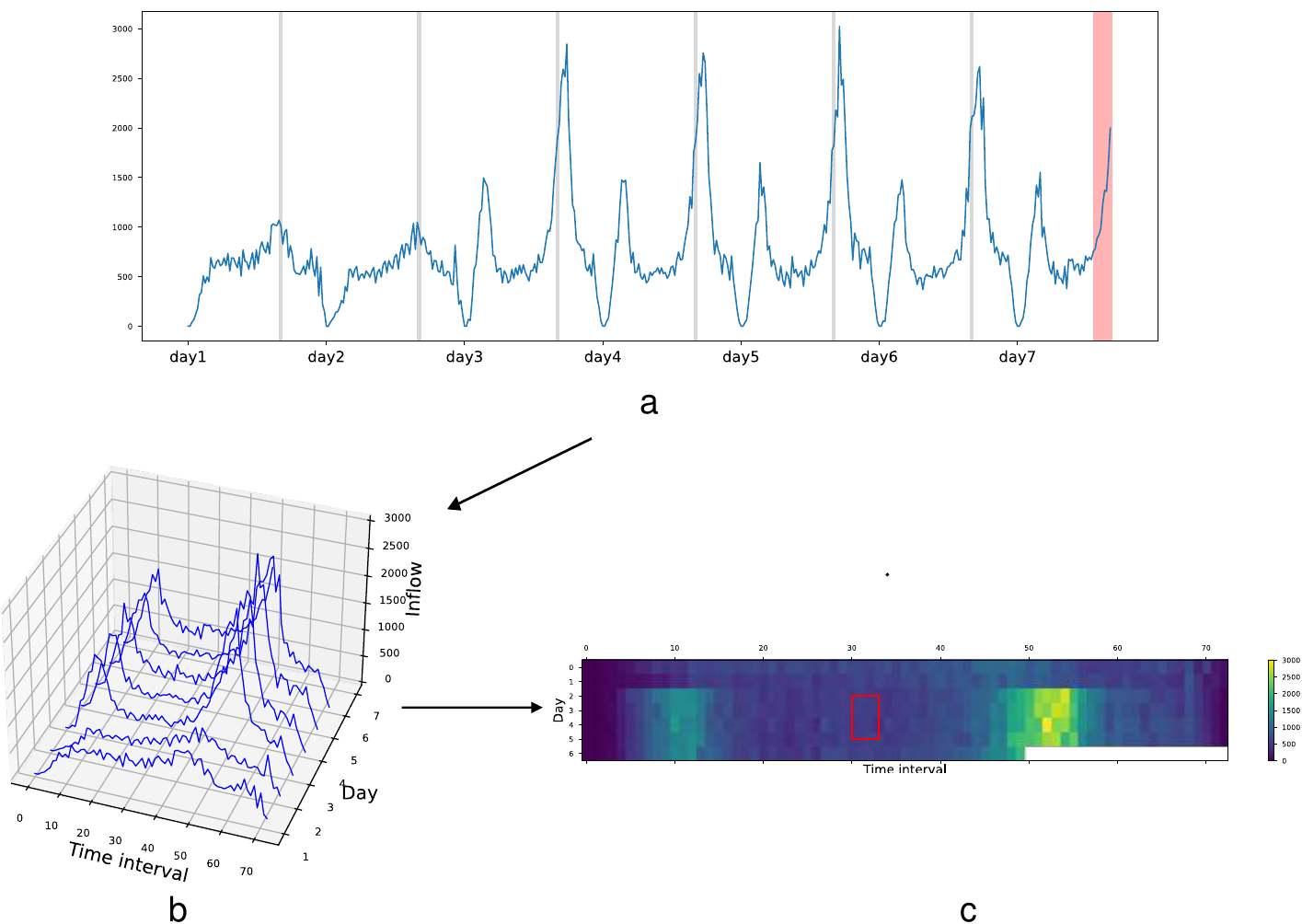}
    \caption{Analysis process of multi-period data. a) raw inflow data of a station in 7 days.
b) folding the inflow data according to the period. c) folding data as a matrix showing structural features similar to photos.
}
    \label{fig2}
\end{figure}

Considering only several past intervals of the same day and the same time interval of several previous days, however, is not sufficient. The passenger flow in the surrounding time intervals of the previous days may also provide valuable insights for predicting passenger flow. To illustrate this point, let's consider the commuting patterns of an individual on a daily basis. For instance, on some workdays, the individual may enter a certain subway station at 8:15 AM, while on other workdays, the individual might arrive at the subway station at 8:45 AM due to various factors such as waking up later or having different work schedules. However, in most cases, the individual tends to enter the station between 8:00 AM and 9:00 AM for his/her morning commute. 

Therefore, we can generalize the fluctuation patterns of each individual's daily commute to the daily fluctuation patterns of passenger flow at each subway station. As a result, the passenger flow between 8:30 AM and 8:45 AM at a specific subway station may exhibit fluctuations. Some days may experience high traffic flow during this time interval, while on other days, it might be lower. However, if we consider the passenger flow between 8 AM and 9 AM as a whole, the fluctuation rate may be smaller, and the stability of passenger flow between 8 AM and 9 AM is informative to the passenger flow prediction between 8:30 and 8:45. Unfortunately, the methods proposed by the aforementioned literature \cite{Li2022ImprovingSB,Guo2019AttentionBS} do not incorporate this information, because the passenger flow from 8:00 AM to 8:30 AM from other days are not considered when predicting the passenger flow from 8:30 AM to 8:45 AM.

To address this, we propose a method that folds the long time series data based on its periodicity and organizes the 1-dimensional time series data into a 2-dimensional matrix, where the passenger flow data in the same days and time intervals can be organized in the same rows and columns. For example, if there are sixty time-intervals in a day and we have data from the past 7 days, we can construct a matrix of size 7 rows by 60 columns. Each row represents the passenger flow for a specific day, while each column represents the passenger flow at the same time interval across different days, as indicated in Fig.\ref{fig2}. By doing this, we can observe that there is correlation between adjacent data points in the matrix. Data points in the adjacent rows represent passenger flow for adjacent time intervals, while data points in the adjacent columns represent passenger flow for adjacent days. Combining the data of adjacent rows and columns allows us to capture the inter and intra periods dependencies, which can be utilized to predict future passenger flow better based on the analysis we conducted earlier. Therefore, a method to integrate information of adjacent points is needed. Coincidentally, in photographs, adjacent points are usually correlated too. Therefore, some image analysis models, such as CNN, are appropriate to the problem because they attempt to extract correlations between adjacent points as much as possible \cite{wu2023timesnet}. Consequently, we consider utilizing a CNN to uncover the characteristics of passenger flow between different periods and employing a GNN to exploit spatial relationship between stations.

In summary, our model consists of three main components. Firstly, for each station, we utilize a CNN to extract information from the historical data, specifically focusing on different periods. Subsequently, the information extracted from each station is fed into a GNN to integrate the information from different stations. Finally, the integrated information from the GNN is fed into a liner layer to generate the final prediction results.

\section{Literature Review}
The prediction of future passenger flow in subway stations is a unique task in spatial-temporal forecasting. Within the domain of traffic, spatial-temporal prediction entails tasks such as forecasting future bike or taxi usage in various city areas and predicting traffic flow at different checkpoints along city roads (and we will use traffic units to represent subway stations, city areas or checkpoints for convenient in following part). These tasks all involve the integration of both spatial and temporal information. In the following section, we will collectively review the literature on temporal-spatial prediction in traffic areas.

In the realm of spatiotemporal data mining, the concern lies in how to extract distinctive features from the spatial and temporal dimensions and the integration method of spatial and temporal information.
\subsection{Spatial information extraction method}

Typically, when exploring spatial features, graph convolutional neural networks (GCNs) are employed, with some studies even utilizing specialized variants such as diffusion graph convolutional neural networks \cite{li2018diffusion}. As the Transformer model has demonstrated remarkable success in the domains of computer vision and natural language processing, attention-based approaches have been employed to mine spatial features in some articles \cite{Xu2020SpatialTemporalTN}. Additionally, some studies consider adopting adaptive graphs, whereby the spatial graph structure is learned from the data, rather than solely relying on spatial connectivity relationships. Furthermore, some articles argue that spatial relationships are inherently time-varying. Hence, these articles learn distinct spatial correlations at different time and incorporate them as inputs to the neural network \cite{Wu2019GraphWF,Feng2022AdaptiveGS}. 

\subsection{Temporal information extraction method}

Regarding the exploration of temporal features, the most straightforward method is to utilize recurrent neural networks (RNNs), such as Long-Short Term Memory (LSTM) and Gated Recurrent Unit (GRU), to capture temporal dependencies \cite{Liu2020PhysicalVirtualCM}. However, some articles contend that RNNs struggle with lengthy time series data. Consequently, they may resort to one-dimensional convolution along the temporal axis to extract temporal features \cite{Yu2017SpatioTemporalGC}. Similarly, with the recent advancements in attention mechanisms, certain articles apply attention mechanisms to the temporal dimension. Therefore, RNNs incorporated with attention mechanisms or Transformer models are employed to handle temporal features \cite{Lin2020PreservingDA}.

\subsection{Integration method of spatial and temporal information}
 Generally, methods for spatial-temporal prediction can be classified into two categories based on how spatial and temporal information are combined. The first category involves extracting spatial information at different time intervals and then integrating the spatial information from these intervals. The second category is characterized by the simultaneous integration of spatial and temporal information.
\begin{figure}[!h]
    \centering
    \includegraphics{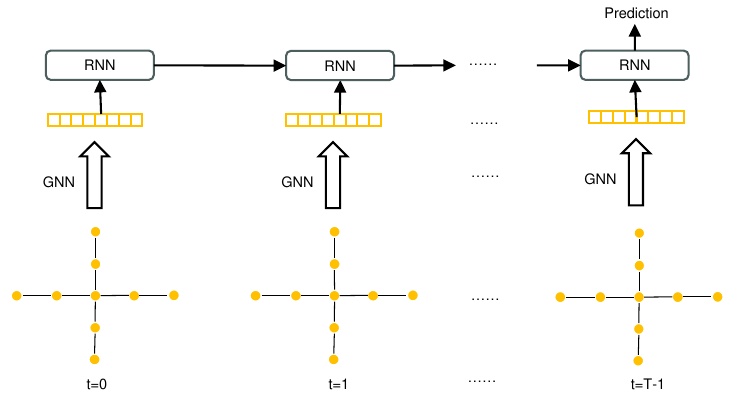}
    \caption{The workflow of the first category of methods. Each node in the bottommost column represents a traffic unit, such as a subway station or a city area. The nodes and edges can be considered as a graph, which reflects the spatial information of the traffic network.}
    \label{fig3}
\end{figure}

To elaborate, in the first category, as depicted in Fig.\ref{fig3}, GNNs are often used to extract spatial information specific to each time interval. The extracted information is then fed into a RNN to integrate the temporal information and generate predictions \cite{Liu2020PhysicalVirtualCM}. We denominate this approach as the "spatial first" strategy to elucidate the preeminence of spatial feature extraction for each time interval, followed by their subsequent fusion along the temporal axis.
\begin{figure}[!h]
    \centering
    \includegraphics[width=\textwidth,height=0.6\textwidth]{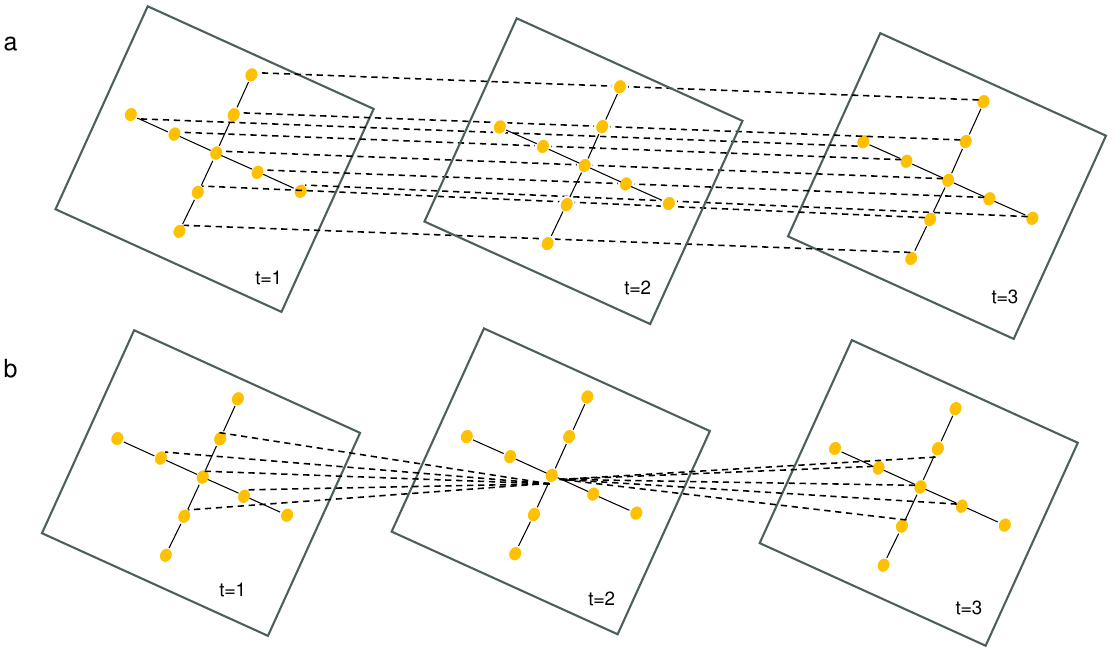}
    \caption{Different Spatial-temporal Graphs. a) represents graph with connections between the same node at adjacent time intervals. b) represents graph with connections not only between the same node at adjacent time intervals but also between adjacent nodes at adjacent time intervals. To ensure clarity in the visualization, only the cross-time connections of the central node and in time 2 are depicted.}
    \label{fig4}
\end{figure}

In the second category of methods, the integration of traffic flow across different time intervals is achieved by establishing connections between the same node in adjacent time intervals, rather than treating graphs from different time intervals as separate graphs. This results in the formation of a unified spatial-temporal graph \cite{Song2020SpatialTemporalSG}, as illustrated in Fig.\ref{fig4}a. In this figure, the dashed lines represent edges connecting the same node in adjacent time intervals, while the solid lines depict edges within the same time period, representing spatial relationships. Within the second category of methods, a common approach involves the utilization of GNNs at each time interval and one-dimensional convolution along the time axis defined by the dashed lines. By iteratively applying GNN and convolutional operations along the spatial and temporal axes, both temporal and spatial information can be concurrently integrated. Some studies have pointed out that simply adding links between the same node from adjacent time intervals is insufficient to comprehensively capture temporal information. This is because there are significant correlations among different nodes across various time intervals \cite{Song2020SpatialTemporalSG}. Therefore, connections between adjacent nodes in adjacent time intervals are also added. This can be represented as Fig.\ref{fig4}b. There are also approaches that employ adaptive and learnable spatiotemporal graphs 
 \cite{Chen2022TAMPS2GCNetsCT}. We name this type of method 'concurrently' to signify that these methods simultaneously explore both temporal and spatial features. 

The previous literature is summarized into the Tab.\ref{tab1} based on the methods for feature extraction method along the temporal axis and spatial axis, and integration method of the temporal and spatial information (Attn. means using attention mechanism and Ada. graph means using adaptive graph in Tab.\ref{tab1}).

\begin{tiny}
\begin{longtable}{ccccccccccccc}
\caption{Summary of existing spatiotemporal prediction models}\label{tab1} \\
\hline
\multirow{2}{*}{Model}                  & \multirow{2}{*}{Conference/ Journal} & \multirow{2}{*}{Year} & \multicolumn{4}{c}{Spatial Axis}                                        

& \multicolumn{3}{c}{Temporal axis} 

& \multicolumn{3}{c}{Spatiotemporal integration mode} \\ 
                          &                     &      & GNN & CNN & Attn. & \begin{tabular}[c]{@{}l@{}}Ada.\\ graph\end{tabular} & RNN    & Convolution    & Attn.   & Spatial first      & Concurrently      & Other      \\ \hline
ST-ResNet\cite{Zhang2016DeepSR}         & AAAI                & 2017 &     & \Checkmark   &       &                                                      &        & \Checkmark             &         & \Checkmark                  &                   &            \\DMVST-Net \cite{Yao2018DeepMS}         & AAAI                & 2018 & \Checkmark   &     &       &                                                      & \Checkmark      &                &         & \Checkmark                  &                   &            \\
DCRNN\cite{li2018diffusion}              & ICLR                & 2018 & \Checkmark   &     &       &                                                      & \Checkmark      &                &         & \Checkmark                  &                   &            \\
STGCN\cite{Yu2017SpatioTemporalGC}              & IJCAI               & 2018 & \Checkmark   &     &       &                                                      &        & \Checkmark              &         &                    & \Checkmark                 &            \\
T-GCN\cite{Zhao2020TGCNAT}           & IEEE TITS           & 2019 & \Checkmark   &     &       &                                                      & \Checkmark      &                &         & \Checkmark                  &                   &            \\
STDN\cite{Yao2018RevisitingSS}             & AAAI                & 2019 &     & \Checkmark   &       &                                                      & \Checkmark      &                & \Checkmark       & \Checkmark                  &                   &            \\
ASTGCN\cite{Guo2019AttentionBS}           & AAAI                & 2019 & \Checkmark   &     & \Checkmark     &                                                      &        & \Checkmark              & \Checkmark       &                    & \Checkmark                 &            \\
Graph WaveNet\cite{Wu2019GraphWF}     & IJCAI               & 2019 & \Checkmark   &     &       & \Checkmark                                                    &        & \Checkmark              &         &                    & \Checkmark                 &            \\
STG2Seq\cite{Bai2019STG2SeqSG}         & IJCAI               & 2019 & \Checkmark   &     &       &                                                      & \Checkmark      &                &         &                    & \Checkmark                 &            \\
ST-MetaNet\cite{Pan2019UrbanTP}      & KDD                 & 2019 & \Checkmark   &     &       & \Checkmark                                                    & \Checkmark      &                &         &                    & \Checkmark                 &            \\
Parallel CNN-LSTM\cite{Ma2019ParallelAO} & IEEE TITS           & 2019 &     & \Checkmark   &       &                                                      & \Checkmark      &                &         & \Checkmark                  &                   &            \\
DSAN\cite{Lin2020PreservingDA}            & KDD                 & 2020 &     &     & \Checkmark     & \Checkmark                                                    &        &                & \Checkmark       &                    & \Checkmark                 &            \\
Conv-GCN\cite{Zhang2020MultigraphCN}          & IET ITS             & 2020 & \Checkmark   &     &       &                                                      &        & \Checkmark              &         &                    & \Checkmark                 &            \\
AGCRN\cite{Bai2020AdaptiveGC}           & NIPS                & 2020 & \Checkmark   &     &       & \Checkmark                                                    & \Checkmark      &                &         & \Checkmark                  &                   &            \\
STSGCN\cite{Song2020SpatialTemporalSG}             & AAAI                & 2020 & \Checkmark   &     &       &                                                      & \Checkmark      &                &         &                    & \Checkmark                 &            \\
SLCNN\cite{Zhang2020SpatioTemporalGS}             & AAAI                & 2020 &     & \Checkmark   &       & \Checkmark                                                    &        & \Checkmark              &         &                    & \Checkmark                 &            \\
STTN\cite{Xu2020SpatialTemporalTN}               & Arxiv               & 2020 & \Checkmark   &     & \Checkmark     &                                                      &        &                & \Checkmark       &                    & \Checkmark                 &            \\

DST-ICRL\cite{Du2020DeepIC}          & IEEE TITS           & 2020 &     & \Checkmark   &       & \Checkmark                                                    & \Checkmark      &                &         & \Checkmark                  &                   &            \\
STGODE\cite{Fang2021SpatialTemporalGO}           & KDD                 & 2021 & \Checkmark   &     &       &                                                      &        & \Checkmark              &         &                    & \Checkmark                 &            \\
STFGNN\cite{Li2020SpatialTemporalFG}           & AAAI                & 2021 & \Checkmark   &     &       &                                                      &        & \Checkmark              &         &                    & \Checkmark                 &            \\
CCRNN\cite{Ye2020CoupledLG}             & AAAI                & 2021 & \Checkmark   &     &       & \Checkmark                                                    & \Checkmark      &                &         & \Checkmark                  &                   &            \\
OGCRNN\cite{Guo2021OptimizedGC}            & IEEE TITS           & 2021 & \Checkmark   &     &       & \Checkmark                                                    &        & \Checkmark              &         &                    & \Checkmark                 &            \\
ResLSTM\cite{Zhang2019DeepLA}           & IEEE TITS           & 2021 & \Checkmark   &     &       &                                                      & \Checkmark      &                & \Checkmark       & \Checkmark                  &                   &            \\
ToGCN\cite{Qiu2021TopologicalGC}             & IEEE TITS           & 2021 & \Checkmark   &     &       & \Checkmark                                                    & \Checkmark      &                &         & \Checkmark                  &                   &            \\
ASTTN\cite{Feng2022AdaptiveGS}           & CIKM                & 2022 &     &     & \Checkmark     & \Checkmark                                                    &        &                & \Checkmark       &                    & \Checkmark                 &            \\

ST-RCNet-knn\cite{Zhang2022ForecastingTC}      & TR part C           & 2022 & \Checkmark   &     &       &                                                      & \Checkmark      &                &         &                    &                   & \Checkmark          \\
GCN-SBULSTM\cite{Chen2022AGC}       & IEEE TITS           & 2022 & \Checkmark   &     &       &                                                      & \Checkmark      &                &         &                    &                   & \Checkmark          \\
PVCGN\cite{Liu2020PhysicalVirtualCM}             & IEEE TITS           & 2022 & \Checkmark   &     &       &                                                      & \Checkmark      &                &         & \Checkmark                  &                   &            \\
HSTGCNT\cite{Huo2023HierarchicalSG}          & IEEE TITS           & 2023 & \Checkmark   &     &       &                                                      &        & \Checkmark              & \Checkmark       &                    & \Checkmark                 &            \\
MVSTGN\cite{Yao2023MVSTGNAM}            & IEEE TMC            & 2023 &     & \Checkmark   & \Checkmark     &                                                      &        &                & \Checkmark       &                    & \Checkmark                 &            \\
IG-Net\cite{Li2023IGNetAI}            & IEEE TITS           & 2023 & \Checkmark   &     &       &                                                      &        &                &         & \Checkmark                  &                   &            \\
IrConv-LSTM\cite{Li2022ImprovingSB}        & TR part C           & 2023 &     & \Checkmark   &       & \Checkmark                                                    &        &                &         & \Checkmark                  &                   &            \\
ST-SSL\cite{ji2023spatio}            & AAAI                & 2023 & \Checkmark   &     &       &                                                      &        & \Checkmark              &         &                    & \Checkmark                 &            \\
MegaCRN\cite{jiang_spatio-temporal_2023}          & AAAI                & 2023 & \Checkmark   &     &       & \Checkmark                                                    & \Checkmark      &                &         &                    & \Checkmark                 &            \\  \hline \\

\end{longtable}
\end{tiny}

There are also methods except deep learning for traffic flow prediction, for example Dynamic Factor Model \cite{Noursalehi2018RealTT} and Dynamic Mode Decomposition \cite{Cheng2021RealTimeFO}. These models often argue that the traffic flow fluctuations of all units are the compound of some prototype modes. Therefore, these models aim to find the prototype modes, prototype modes weights of traffic units and the transmit matrix of the modes weights. These models can decrease the number of parameters and obtain more robust prediction.
\section{Method}
We will begin by providing an overview of our proposed method. We approach the data analysis within the context of a single spatiotemporal graph. Unlike previous research, where each node represents passenger flow for a specific station and time interval, we consider each node to represent the aggregated passenger flow for all past time intervals at a particular station. Consequently, the feature associated with each node represents a long time series of data.

To explore the inter and intra-period correlations in the historical passenger flow, we reshape the time series data into a matrix, where each row represents a specific period. We then employ a CNN to extract features from the matrix of historical passenger flow data for each node. Additionally, we concatenate weather information with the features extracted by the CNN.

Next, we utilize a GNN to incorporate the feature information from neighboring stations, thereby capturing the spatial relationships within the subway network.

\begin{figure}[!h]
    \centering
    \includegraphics{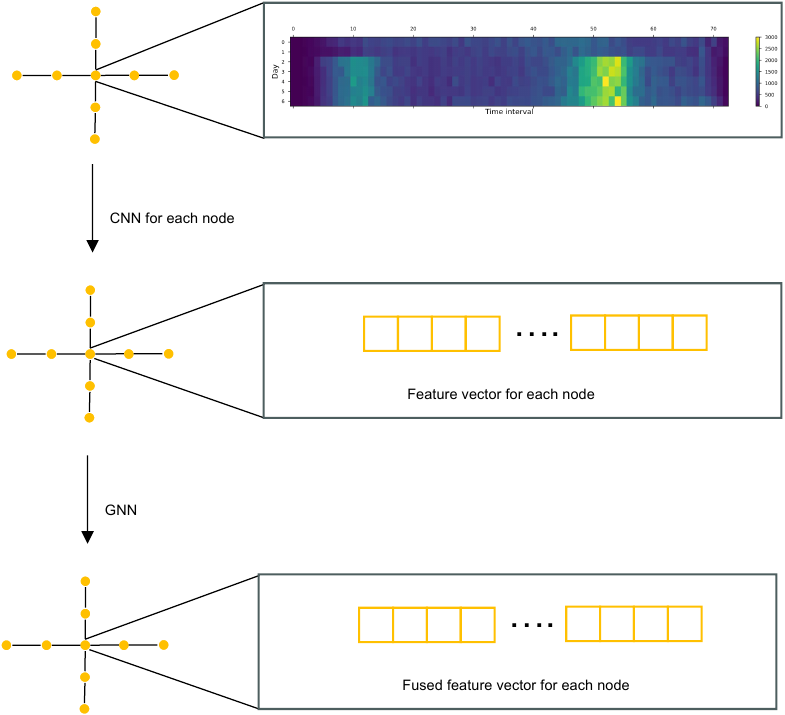}
    \caption{Workflow of our appeared methods. The left part represents the graph and the right part of the figure represent the note feature.}
    \label{fig5}
\end{figure}

Finally, we employ some liner layers to predict passenger flow for each subway station. The workflow of our proposed method is illustrated as Fig.\ref{fig5} (The final liner layers are obviated). 
\subsection{Data Preprocessing}
For each station, we start by extracting passenger flow data from multiple historical periods. This data is then transformed into matrices based on the corresponding periods. Since both the historical inflow and outflow data are available, two separate matrices are obtained for each station. As a result, the input data dimensionality for each station becomes the product of the period length, the number of periods, and two (indicating inflow and outflow). 

For example, if the passenger flow in 8:00AM to 9:00AM on July 15 need to be predicted, we will extract the passenger flow between 8:00AM July 1 and 8:00AM July 15 from the history record. Then the history passenger flow for each station will be reshaped as a matrix.
\subsection{CNN Module}
After the historical passenger flow data of each station has been transformed into matrix form, features from different periods are extracted using a CNN model. In this section, the definition of the convolution operator will be presented, and an overview of the framework of the employed CNN model will be provided.

Convolutional operations are mathematical operations that involve the application of a filter (also known as a kernel) to an input data. The convolution operation computes the element-wise multiplication of the filter and the corresponding elements of the input data, followed by the summation of these products to produce a single output value. By sliding the filter over the entire input data, the convolution operation generates a new matrix, known as a feature map, that summarizes the important features present in the input.

Assuming the input data is a two-dimensional matrix of size $M$ $\times$ $N$, represented as $I$, and the filter is a two-dimensional matrix of size $K_m$ $\times$ $K_n$, represented as $K$, the convolution operation can be expressed as Eq.\ref{eq1}:
\begin{equation}
    C_{i,j}=\sum_{n}\sum_{m}\left(K_{m,n}\times I_{i-m,j-n}\right)
    \label{eq1}
\end{equation}
Here, $C_{i,j}$represents the element of the resulting output of the convolution operation at position $(i, j)$. $I_{i-m,j-n}$denotes the pixel value of the input data matrix at position\ $(i-m,\ j-n)$, and $K_{m,n}$ represents the weight at position $(m,\ n)$ of the filter.

Because adjacent rows and columns represent adjacent days and time intervals in our data matrix, consequently, the convolutional operator can effectively integrate information from neighboring days and time intervals. This integration enables a more comprehensive understanding of the patterns within the passenger flow data.

A typical CNN model consists of several convolutional layers, pooling layers, fully connected layers, and activation functions. Among them, the convolutional layers are the most critical components. Each convolutional layer typically contains multiple learnable convolutional kernels, which are utilized to extract diverse types of features from the input data. 

In our study, the CNN model employed consists of two 2D convolution layers and two pooling layers.These two convolutional layers each contain 32 and 64 learnable 3x3 two-dimensional convolutional kernels, respectively. After obtaining the feature output from the CNN, it is concatenated with weather information, specifically indicating whether it is rain on the predicted day. 

Despite its simplicity, the model showcases exceptional performance, achieving state-of-the-art accuracy. This success can be attributed to the convolutional operations' capacity to capture the inherent patterns present in the passenger flow data.
\subsection{GNN Module}
After obtaining features for all stations using the CNN model, a GNN is employed to account for the interconnections between different stations. In this paper, we consider the simplest GNN architecture that performs message passing in the spatial domain. 

The graph neural network for message passing in a graph can be described by the following procedure. If we have $n$ nodes, and each node has a feature vector of dimension $p$, we can represent our data as an $n\times p$ matrix $X$. Furthermore, if we have knowledge of the adjacency matrix $A$, which is an $n\times n$ matrix representing the interconnections between the $n$ nodes. A GNN can be performed as the following steps:

1 Compute the message propagation: Using the current node feature matrix $X^{\left(l\right)}$ and the normalized adjacency matrix $\hat{A}$, the messages between nodes are calculated as Eq.\ref{eq2}:
\begin{equation}
    M^{\left(l+1\right)}=\sigma\left(\hat{A}\cdot X^{\left(l\right)}\cdot W^{\left(l\right)}\right)
    \label{eq2}
\end{equation}
Here, $\sigma$ denotes the activation function, $X^{\left(l\right)}$ represents the node feature matrix at layer $l$, and $W^{\left(l\right)}$ is the weight matrix for layer $l$.

2 Update the node features: the node feature matrix for the next layer is directly updated with the messages as Eq.\ref{eq3} indicates. By doing this, the node features are immediately influenced by the propagated information, allowing for capturing the spatial dependencies between stations.
\begin{equation}
    X^{\left(l+1\right)}=M^{\left(l+1\right)}
    \label{eq3}
\end{equation}

3 Repeat the process: The above steps are repeated for each layer of the GNN, with each layer having its own set of weight matrices and performing message passing and feature update accordingly.

Through this iterative process, the GNN leverages the interconnections between nodes to propagate and integrate information across the graph, gradually improving the node representations. By effectively capturing the spatial relationships and considering the graph structure, the GNN provides a mechanism to utilize the dependencies between different stations.

\subsection{Prediction Head}
The output features from the GNN are ultimately concatenated with the output features from the previous CNN, similar to the residual connection approach \cite{He2015DeepRL}. This yields a final representation of features for each station. Subsequently, an MLP consist of 2 linear layers is employed to predict the future passenger flow for each station at intervals of 15 minutes, 30 minutes, 45 minutes, and one hour, utilizing the final features of each station as input.
\section{Experiment}
\subsection{Data}
The data utilized in our experiments comprises the subway passenger flow data of Shanghai City in the year 2016, from July to September \cite{Wang2023TowardsEA}. And there were 288 stations in Shanghai subway system at that time. The dataset used in our experiment spans a total of 92 days, with passenger flow data recorded at fifteen-minute intervals. This results in a total of 73 time intervals in a single day. Our focus is on predicting the inflow and outflow numbers for every subway station in the following hour.

Since the passenger flow data during the Mid-Autumn Festival was considered as out-of-distribution sample, we excluded the corresponding week containing the festival from our dataset. Consequently, the total number of days in our data amounted to 85. We designated the initial two months' data as the training set and the subsequent week as the validation set, while the remaining data served as the test set.
\subsection{Implement details}
To implement our prediction model, the Python package PyTorch was utilized, and the learning rate was initialized to 0.001 and progressively decreased by 0.05 every two epochs. The model was trained for 200 epochs. As part of our experimental setup, a hyperparameter grid search was conducted and the search space is depicted in Tab.\ref{tab2}. 

The model with the best performance on the validation set was selected for testing.
\begin{table}[]
\centering
\caption{Hyperparameters and search space}
\begin{tabular}{cc}
\toprule
Hyperparameters               & Search   space    \\
\midrule
Number   of GNN layers        & {[}1,2,4{]}        \\
Batch   size                  & {[}4,8,16{]}       \\
Weather   embedding dimension & {[}2,4{]}          \\
Feature   dimension in GNN    & {[}64,128,256{]}   \\
\bottomrule
\end{tabular}
\label{tab2}
\end{table}
In our experiment, the input period was set to 14, representing a duration of 14 days. The loss function employed was the L1-norm loss.
\subsection{Evaluation}
To evaluate the results of our prediction, we employed the Mean Absolute Error (MAE) and Root Mean Squared Error (RMSE) metrics. The formulations for MAE and RMSE are defined as Eq.\ref{eq4} and Eq.\ref{eq5}:

\begin{equation}
      MAE=\frac{1}{n}\sum_{i=1}^{n}\left|y_{pred_i}-y_{actual_i}\right|
      \label{eq4}
\end{equation}
\begin{equation}
   RMSE=\sqrt{\frac{1}{n}\sum_{i=1}^{n}\left(y_{pred_i}-y_{actual_i}\right)^2}
\label{eq5}
\end{equation}

In these expressions, $y_{pred_i}$ represents the predicted passenger flow values of station $i$ , $y_{actual_i}$ represents the actual passenger flow values of station $i$, and $n$ represents the total number of stations. 
\subsection{Benchmark Models}
The benchmark models in our paper can be categorized into three classes: 1) Traditional deep learning algorithms: GRU and LSTM; 2) State-of-the-art models for traffic demand prediction and traffic state prediction: STGCN, AGCRN, STTN, DCRNN, CCRNN; 3) Open code models specifically designed for subway passenger flow prediction: ConvGCN and PVGCN. A brief description of the benchmark methods is provided below.

STGCN \cite{Yu2017SpatioTemporalGC} (Spatiotemporal Graph Convolutional Networks): It uses GNN to capture spatial features and employs convolutional operators in the time axis to obtain temporal features. The model repeats this process iteratively to generate the final prediction.

AGCRN \cite{Bai2020AdaptiveGC} (Adaptive Graph Convolutional Recurrent Network): Presented at the 2020 NIPS conference. The main innovation of this model is the use of adaptive graph structures learned from the data to consider the spatial relationships of different nodes at different time steps.

STTN \cite{Xu2020SpatialTemporalTN} (Spatial-Temporal Transformer Networks): This model employs a spatiotemporal transformer network to predict traffic flow in future times.

DCRNN \cite{li2018diffusion} (Diffusion Convolutional Recurrent Neural Network): Introduced at the ICRL 2018 conference, this model constructs a diffusion convolutional neural network on a directed graph based on the theory of random walk and uses GRU to integrate information from different time periods.

CCRNN \cite{Ye2020CoupledLG} (Coupled Layer-wise Convolutional Recurrency Neural Network): Presented at the 2021 AAAI conference, this model considers spatial information by learning adaptive hierarchical adjacency matrices. It integrates spatial information from different time steps using an RNN.

ConvGCN \cite{Zhang2020MultigraphCN}: This model is specifically designed for short-term subway passenger flow prediction. It combines GNN with three-dimensional convolutional neural networks (3D CNN). The 3D CNN is applied innovatively to integrate the information of station inflow and outflow, as well as to capture the correlation of flow patterns and the high-level correlation between stations at different distances.

PVGCN \cite{Liu2020PhysicalVirtualCM} (Physical-Virtual Collaboration Graph Network): This model is specifically designed for predicting passenger flow in subway systems. It derives three different graphs to consider the relationships between stations from three perspectives. The model combines GCN with GRU to jointly consider spatiotemporal features.

The implementation of PVGCN was based on the code offered by the author, while the implementation of other methods was based on the repository libcity \cite{Wang2023TowardsEA}.
\subsection{Result}
In Tab.\ref{tab3}, we present the final prediction errors of all the models evaluated in our experiments.
\begin{table}
\caption{Result of all benchmark models and our model}
\label{tab3}
\resizebox{\textwidth}{20mm}{

\begin{tabular}{cccccccccccc}
\toprule
Time  & Metrics & GRU    & LSTM   & STGCN  & AGCRN  & STTN   & DCRNN  & CCRNN  & ConvGCN & PVGCN & MPSTN           \\
\midrule
\multirow{2}{*}{15min} & MAE     & 29.403 & 32.273 & 30.838 & 24.209 & 30.307 & 28.000 & 25.533 & 28.476  &  22.627 & \textbf{20.724} \\
      & RMSE    & 63.230 & 68.496 & 65.710 & 47.549 & 62.109 & 53.946 & 46.805 & 61.122  &41.264 & \textbf{36.856} \\
\multirow{2}{*}{30min} & MAE     &  29.486 & 33.280 & 31.224 & 25.312 & 30.460 & 31.847 & 26.231 & 30.267  &23.159 & \textbf{21.082} \\
      & RMSE    & 63.578 & 71.453 & 68.027 & 51.397 & 62.084 & 62.999 & 49.353 & 65.556  & 43.107& \textbf{37.601} \\
\multirow{2}{*}{45min} & MAE     & 29.780 & 33.835 & 31.737 & 26.754 & 31.524 & 36.703 & 27.660 & 31.965  &23.902& \textbf{21.361} \\
      & RMSE    &  64.271 & 74.099 & 69.789 & 55.295 & 65.305 & 77.398 & 53.481 & 68.764  & 45.283 & \textbf{38.317} \\
\multirow{2}{*}{60min} & MAE     & 30.226 & 35.312 & 32.774 & 27.923 & 32.609 & 41.223 & 29.640 & 34.317  & 24.477 & \textbf{21.521} \\
      & RMSE     & 65.094& 79.639 & 72.377 & 58.619 & 68.134 & 91.825 & 58.881 & 75.068  & 47.333 & \textbf{38.743}
      \\ \bottomrule
\end{tabular}}
\end{table}
It can be observed that our method consistently achieves lower errors, both in terms of MAE and RMSE, compared to the benchmark methods. Specifically, for the task of predicting the passenger flow in and out of stations for the next 15 minutes, our method is the only one that reduces the MAE below 21. 

Moreover, as the prediction horizon extends to 30 minutes, 45 minutes, and 60 minutes, our method exhibits a substantially slight error increase relative to other methods. In contrast, the other methods demonstrate a notable increase in prediction errors as the forecast horizon lengthens, especially for the prediction of passenger flow for the next 60 minutes. In this case, our method achieves a prediction error of only 21.521 in MAE, whereas the best-performing method among the others attains an MAE of approximately 25. This implies that our method enables earlier and more accurate prediction of different passenger flow variations, providing subway station management personnel more time to employ precise measures in response to varying passenger demands.

The reason for these findings is that our method considers not only the data from previous time steps but also incorporates historical passenger flow data from the corresponding time intervals on previous days. By incorporating this additional temporal context, our method effectively captures complex patterns in the passenger flow data, leading to improved prediction accuracy.
\subsection{Ablation experiment}
To assess the effectiveness of different components in our model, several ablation experiments were conducted. Firstly, the module incorporating weather information was eliminated. Secondly, the CNN module was removed. Consequently, predictions were made for each station independently using CNNs, without accounting for the spatial correlations among stations. The result is provided in Tab.\ref{tab4}

\begin{table}[]
\centering
\caption{Result of ablation experiment}
\begin{tabular}{ccccc}
\\ \toprule
Time  & Metrics & Without GNN & Without Weather & Original model  \\ \midrule
\multirow{2}{*}{15min} & MAE     & 21.493      & 21.270          & \textbf{20.724} \\
      & RMSE    & 38.290      & 38.482          & \textbf{36.856} \\
\multirow{2}{*}{30min} & MAE     & 21.792      & 21.511          & \textbf{21.082} \\
      & RMSE    & 42.217      & 40.873          & \textbf{37.601} \\
\multirow{2}{*}{45min} & MAE     & 21.861      & 21.705          & \textbf{21.361} \\
      & RMSE    & 42.784      & 41.125          & \textbf{38.317} \\
\multirow{2}{*}{60min} & MAE     & 22.252      & 21.762          & \textbf{21.521} \\
      & RMSE     & 44.034      & 42.023          & \textbf{38.743} \\ \bottomrule
\end{tabular}
\label{tab4}
\end{table}

From the ablation experiments, it can be observed that when the weather module was removed, a slight decrease in the prediction accuracy occurred, indicating the effect of weather conditions in the forecast of subway passenger flow.

Moreover, a slight decrease in prediction accuracy can be observed when the GNN module was removed. However, the impact is not substantial. It is suggested that the prediction accuracy is not heavily influenced by spatial correlations among stations. Four explanations can be considered for this phenomenon.

Firstly, subway passenger flow prediction might differ from other types of traffic prediction. For instance, in predicting the traffic flow of different detectors in a specific region, the data values between adjacent detectors are highly correlated. However, it may not hold true for subway passenger flow since individuals using the subway system might not change their behavior based on the actions of passengers at other stations. In other words, the ridership at a particular station might be relatively independent. This distinction sets apart subway passenger flow prediction from other traffic state predictions.

Secondly, spatial correlations might not only exist in the same time interval but could also extend across different time intervals. For example, if there is significant exchange of passengers between two stations, but it takes 30 minutes to travel between them, it means that a large influx of passengers at the origin station will result in a substantial outflow of passengers at the destination station after 30 minutes. Thus, the relevant spatial connections exist across different moments in time. However, effectively considering such relationships remains a challenge for current GNN approaches.

Furthermore, the spatial relationship of subway station ridership is highly complex and cannot be adequately represented solely by the adjacency matrix based on the physical proximity of the stations \cite{Liu2020PhysicalVirtualCM}. It is possible that there is a certain relationship between the passenger flows of physically adjacent stations, but there are also correlations among passenger flows of stations that are not physically adjacent.

Another possible reason is that spatial relationships are different across different times. For instance, there might be a strong passenger flow correlation between stations A and B at 8 AM, indicating that a high influx of passengers at station A at a particular moment will lead to a significant outflow of passengers at station B after a certain time interval, such as 30 minutes. However, at other times, such as 12 AM or evening hours, the passenger flow correlation between stations A and B might be weak, implying that the inflow at station A and the outflow at station B are not significantly related.

Based on the above analysis, it can be concluded that, firstly, the inflow number at individual subway stations might be relatively independent. Secondly, the spatial relationships among the inflow and outflow of different stations are complex and different across multiple time intervals, making it challenging for traditional GNN to effectively capture these relationships. 
\section{Conclusion and Future Work}
In this paper, the objective is to enhance the accuracy of short-term subway passenger flow prediction by utilizing features both within each period and across periods. To the best of our knowledge, this research is the first to incorporate the ridership of adjacent days and adjacent time intervals for predicting feature passenger flow in subway system.

To better capture the intra-period and inter-period characteristics of subway passenger flow, we propose an innovative approach that folds the passenger flow data over a longer time period according to its periodicity. This transforms the one-dimensional data into a two-dimensional form, enabling the application of image processing techniques, specifically CNNs, to integrate the features within and across periods.

Furthermore, we also consider the integration of spatial features of passenger flow across different subway stations using GNN. We conduct an ablation experiment to demonstrate the impact of incorporating spatial features on the accuracy of passenger flow prediction. Additionally, we investigate the influence of external factors, weather conditions, on the prediction accuracy. Our findings suggest that both spatial relationships and weather factors contribute to improvements in passenger flow prediction accuracy, albeit to a modest extent.

We acknowledge that several aspects could be considered for future work in this field. 

Firstly, the exploitation of spatial relationships among subway stations can be improved. Despite the relatively insignificant impact of spatial relationships on passenger flow prediction accuracy found in our study, as indicated by the ablation experiments, it is possible that more advanced modules capturing spatial relationships effectively, particularly the varying relationships in different time intervals, can improve the prediction accuracy remarkably. We acknowledge that some existing models for short-term traffic volume prediction have made advancements in considering such complex spatial relationships \cite{Zhang2020SpatioTemporalGS}. Therefore, we believe that combining our approach with the methods proposed by previous researchers for capturing intricate spatial relationships could potentially result in improved prediction accuracy.

On the other hand, the consideration of multiple periodicities in subway passenger flow data can be explored. For instance, in addition to daily cycles, weekly cycles could also be taken into account. For example, if we have access to historical data spanning four weeks, with each day divided into 60 time slots, we can fold the historical data twice, resulting in a 3D tensor of size 4x7x60. This approach further extends the folding technique proposed in our paper and captures the features associated with multiple periodicities. With the obtained 3D tensor, 3D CNN can be employed to extract features within and across different periods, thus may enable more accurate passenger flow prediction. However, it need be noted that such an operation may require a larger training dataset to tr ain a 3D CNN effectively. Additionally, it is worth considering that CNNs are no longer the state-of-the-art models in computer vision. For instance, the vision transformer model, which is currently popular in the field of computer vision, often outperforms CNNs. And we have also observed studies  using vision transformers to handle multi-periodicities in time series data \cite{wu2023timesnet}. Therefore, if more data becomes available, employing vision transformers to extract features related to multiple periodicities could be a promising approach.

\section{Author Contribution}
The authors confirm contribution to the paper as follows: study conception and design: Xiannan Huang, Chao Yang; data collection: Xiannan Huang; analysis and interpretation of results:Xiannan Huang, Chao Yang; draft manuscript preparation: Xiannan Huang, Quan Yuan. All authors reviewed the results and approved the final version of the manuscript.

\section{Acknowledgements}
This work was supported by the Research Support Scheme of the Open Society Institute/ Higher Education Program, grant no. 285/97, and through a national Academy of education/ Spencer Postdoctoral Fellowship.

\newpage

\bibliographystyle{trb}
\bibliography{trb_template}

\begin{thebibliography}{40}
\providecommand{\natexlab}[1]{#1}

\bibitem[{Liu et~al.(2022)Liu, Chen, Wu, Zhen, Li, and
  Lin}]{Liu2020PhysicalVirtualCM}
Liu, L., J.~Chen, H.~Wu, J.~Zhen, G.~Li, and L.~Lin, Physical-Virtual
  Collaboration Modeling for Intra- and Inter-Station Metro Ridership
  Prediction. \emph{IEEE Transactions on Intelligent Transportation Systems},
  Vol.~23, 2022, pp. 3377--3391.

\bibitem[{Li et~al.(2023{\natexlab{a}})Li, Xu, Zhang, Shi, Yue, and
  Li}]{Li2022ImprovingSB}
Li, X., Y.~Xu, X.~Zhang, W.~Shi, Y.~Yue, and Q.~Li, Improving short-term bike
  sharing demand forecast through an irregular convolutional neural network.
  \emph{Transportation Research Part C: Emerging Technologies}, Vol. 147,
  2023{\natexlab{a}}, p. 103984.

\bibitem[{Guo et~al.(2019)Guo, Lin, Feng, Song, and Wan}]{Guo2019AttentionBS}
Guo, S., Y.~Lin, N.~Feng, C.~Song, and H.~Wan, Attention Based Spatial-Temporal
  Graph Convolutional Networks for Traffic Flow Forecasting. In \emph{AAAI
  Conference on Artificial Intelligence}, 2019.

\bibitem[{Wu et~al.(2023)Wu, Hu, Liu, Zhou, Wang, and Long}]{wu2023timesnet}
Wu, H., T.~Hu, Y.~Liu, H.~Zhou, J.~Wang, and M.~Long, TimesNet: Temporal
  2D-Variation Modeling for General Time Series Analysis. In \emph{The Eleventh
  International Conference on Learning Representations}, 2023.

\bibitem[{Li et~al.(2018)Li, Yu, Shahabi, and Liu}]{li2018diffusion}
Li, Y., R.~Yu, C.~Shahabi, and Y.~Liu, Diffusion Convolutional Recurrent Neural
  Network: Data-Driven Traffic Forecasting. In \emph{International Conference
  on Learning Representations}, 2018.

\bibitem[{Xu et~al.(2020)Xu, Dai, Liu, Gao, Lin, Qi, and
  Xiong}]{Xu2020SpatialTemporalTN}
Xu, M., W.~Dai, C.~Liu, X.~Gao, W.~Lin, G.-J. Qi, and H.~Xiong,
  Spatial-Temporal Transformer Networks for Traffic Flow Forecasting.
  \emph{ArXiv}, Vol. abs/2001.02908, 2020.

\bibitem[{Wu et~al.(2019)Wu, Pan, Long, Jiang, and Zhang}]{Wu2019GraphWF}
Wu, Z., S.~Pan, G.~Long, J.~Jiang, and C.~Zhang, Graph WaveNet for deep
  spatial-temporal graph modeling. In \emph{Proceedings of the Twenty-Eighth
  International Joint Conference on Artificial Intelligence}, United States of
  America, 2019, pp. 1907--1913.

\bibitem[{Feng and Tassiulas(2022)}]{Feng2022AdaptiveGS}
Feng, A. and L.~Tassiulas, Adaptive Graph Spatial-Temporal Transformer Network
  for Traffic Forecasting. In \emph{Proceedings of the 31st ACM International
  Conference on Information and Knowledge Management}, Association for
  Computing Machinery, New York, NY, USA, 2022, CIKM '22, p. 3933–3937.

\bibitem[{Yu et~al.(2017)Yu, Yin, and Zhu}]{Yu2017SpatioTemporalGC}
Yu, T., H.~Yin, and Z.~Zhu, Spatio-Temporal Graph Convolutional Networks: A
  Deep Learning Framework for Traffic Forecasting. In \emph{International Joint
  Conference on Artificial Intelligence}, 2017.

\bibitem[{Lin et~al.(2020)Lin, Bai, Jia, Yang, and You}]{Lin2020PreservingDA}
Lin, H., R.~Bai, W.~Jia, X.~Yang, and Y.~You, Preserving Dynamic Attention for
  Long-Term Spatial-Temporal Prediction. \emph{Proceedings of the 26th ACM
  SIGKDD International Conference on Knowledge Discovery \& Data Mining}, 2020.

\bibitem[{Song et~al.(2020)Song, Lin, Guo, and Wan}]{Song2020SpatialTemporalSG}
Song, C., Y.~Lin, S.~Guo, and H.~Wan, Spatial-Temporal Synchronous Graph
  Convolutional Networks: A New Framework for Spatial-Temporal Network Data
  Forecasting. In \emph{AAAI Conference on Artificial Intelligence}, 2020.

\bibitem[{Chen et~al.(2022{\natexlab{a}})Chen, Segovia-Dominguez, Coskunuzer,
  and Gel}]{Chen2022TAMPS2GCNetsCT}
Chen, Y., I.~Segovia-Dominguez, B.~Coskunuzer, and Y.~R. Gel, TAMP-S2GCNets:
  Coupling Time-Aware Multipersistence Knowledge Representation with
  Spatio-Supra Graph Convolutional Networks for Time-Series Forecasting. In
  \emph{International Conference on Learning Representations},
  2022{\natexlab{a}}.

\bibitem[{Zhang et~al.(2016)Zhang, Zheng, and Qi}]{Zhang2016DeepSR}
Zhang, J., Y.~Zheng, and D.~Qi, Deep Spatio-Temporal Residual Networks for
  Citywide Crowd Flows Prediction. In \emph{AAAI Conference on Artificial
  Intelligence}, 2016.

\bibitem[{Yao et~al.(2018{\natexlab{a}})Yao, Wu, Ke, Tang, Jia, Lu, Gong, Ye,
  and Li}]{Yao2018DeepMS}
Yao, H., F.~Wu, J.~Ke, X.~Tang, Y.~Jia, S.~Lu, P.~Gong, J.~Ye, and Z.~J. Li,
  Deep Multi-View Spatial-Temporal Network for Taxi Demand Prediction. In
  \emph{AAAI Conference on Artificial Intelligence}, 2018{\natexlab{a}}.

\bibitem[{Zhao et~al.(2020)Zhao, Song, Zhang, Liu, Wang, Lin, Deng, and
  Li}]{Zhao2020TGCNAT}
Zhao, L., Y.~Song, C.~Zhang, Y.~Liu, P.~Wang, T.~Lin, M.~Deng, and H.~Li,
  T-GCN: A Temporal Graph Convolutional Network for Traffic Prediction.
  \emph{IEEE Transactions on Intelligent Transportation Systems}, Vol.~21,
  2020, pp. 3848--3858.

\bibitem[{Yao et~al.(2018{\natexlab{b}})Yao, Tang, Wei, Zheng, and
  Li}]{Yao2018RevisitingSS}
Yao, H., X.~Tang, H.~Wei, G.~Zheng, and Z.~J. Li, Revisiting Spatial-Temporal
  Similarity: A Deep Learning Framework for Traffic Prediction. In \emph{AAAI
  Conference on Artificial Intelligence}, 2018{\natexlab{b}}.

\bibitem[{Bai et~al.(2019)Bai, Yao, Kanhere, Wang, and
  Sheng}]{Bai2019STG2SeqSG}
Bai, L., L.~Yao, S.~S. Kanhere, X.~Wang, and Q.~Sheng, STG2Seq:
  Spatial-temporal Graph to Sequence Model for Multi-step Passenger Demand
  Forecasting. In \emph{International Joint Conference on Artificial
  Intelligence}, 2019.

\bibitem[{Pan et~al.(2019)Pan, Liang, Wang, Yu, Zheng, and
  Zhang}]{Pan2019UrbanTP}
Pan, Z., Y.~Liang, W.~Wang, Y.~Yu, Y.~Zheng, and J.~Zhang, Urban Traffic
  Prediction from Spatio-Temporal Data Using Deep Meta Learning.
  \emph{Proceedings of the 25th ACM SIGKDD International Conference on
  Knowledge Discovery \& Data Mining}, 2019.

\bibitem[{Ma et~al.(2019)Ma, Zhang, Du, Ding, and Sun}]{Ma2019ParallelAO}
Ma, X., J.~Zhang, B.~Du, C.~Ding, and L.~Sun, Parallel Architecture of
  Convolutional Bi-Directional LSTM Neural Networks for Network-Wide Metro
  Ridership Prediction. \emph{IEEE Transactions on Intelligent Transportation
  Systems}, Vol.~20, 2019, pp. 2278--2288.

\bibitem[{Zhang et~al.(2020{\natexlab{a}})Zhang, Chen, and
  Guo}]{Zhang2020MultigraphCN}
Zhang, J., F.~Chen, and Y.~Guo, Multi‐graph convolutional network for
  short‐term passenger flow forecasting in urban rail transit. \emph{IET
  Intelligent Transport Systems}, 2020{\natexlab{a}}.

\bibitem[{BAI et~al.(2020)BAI, Yao, Li, Wang, and Wang}]{Bai2020AdaptiveGC}
BAI, L., L.~Yao, C.~Li, X.~Wang, and C.~Wang, Adaptive Graph Convolutional
  Recurrent Network for Traffic Forecasting. In \emph{Advances in Neural
  Information Processing Systems} (H.~Larochelle, M.~Ranzato, R.~Hadsell,
  M.~Balcan, and H.~Lin, eds.), Curran Associates, Inc., 2020, Vol.~33, pp.
  17804--17815.

\bibitem[{Zhang et~al.(2020{\natexlab{b}})Zhang, Chang, Meng, Xiang, and
  Pan}]{Zhang2020SpatioTemporalGS}
Zhang, Q., J.~Chang, G.~Meng, S.~Xiang, and C.~Pan, Spatio-Temporal Graph
  Structure Learning for Traffic Forecasting. In \emph{AAAI Conference on
  Artificial Intelligence}, 2020{\natexlab{b}}.

\bibitem[{Du et~al.(2020)Du, Peng, Wang, Bhuiyan, Wang, Gong, Liu, and
  Li}]{Du2020DeepIC}
Du, B., H.~Peng, S.~Wang, M.~Z.~A. Bhuiyan, L.~Wang, Q.~Gong, L.~Liu, and
  J.~Li, Deep Irregular Convolutional Residual LSTM for Urban Traffic Passenger
  Flows Prediction. \emph{IEEE Transactions on Intelligent Transportation
  Systems}, Vol.~21, 2020, pp. 972--985.

\bibitem[{Fang et~al.(2021)Fang, Long, Song, and
  Xie}]{Fang2021SpatialTemporalGO}
Fang, Z., Q.~Long, G.~Song, and K.~Xie, Spatial-Temporal Graph ODE Networks for
  Traffic Flow Forecasting. \emph{Proceedings of the 27th ACM SIGKDD Conference
  on Knowledge Discovery \& Data Mining}, 2021.

\bibitem[{Li and Zhu(2020)}]{Li2020SpatialTemporalFG}
Li, M. and Z.~Zhu, Spatial-Temporal Fusion Graph Neural Networks for Traffic
  Flow Forecasting. In \emph{AAAI Conference on Artificial Intelligence}, 2020.

\bibitem[{Ye et~al.(2020)Ye, Sun, Du, Fu, and Xiong}]{Ye2020CoupledLG}
Ye, J., L.~Sun, B.~Du, Y.~Fu, and H.~Xiong, Coupled Layer-wise Graph
  Convolution for Transportation Demand Prediction. In \emph{AAAI Conference on
  Artificial Intelligence}, 2020.

\bibitem[{Guo et~al.(2021)Guo, Hu, Qian, Liu, Zhang, Sun, Gao, and
  Yin}]{Guo2021OptimizedGC}
Guo, K., Y.~Hu, Z.~S. Qian, H.~Liu, K.~Zhang, Y.~Sun, J.~Gao, and B.~Yin,
  Optimized Graph Convolution Recurrent Neural Network for Traffic Prediction.
  \emph{IEEE Transactions on Intelligent Transportation Systems}, Vol.~22,
  2021, pp. 1138--1149.

\bibitem[{Zhang et~al.(2021)Zhang, Chen, Cui, Guo, and Zhu}]{Zhang2019DeepLA}
Zhang, J., F.~Chen, Z.~Cui, Y.~Guo, and Y.~Zhu, Deep Learning Architecture for
  Short-Term Passenger Flow Forecasting in Urban Rail Transit. \emph{IEEE
  Transactions on Intelligent Transportation Systems}, Vol.~22, 2021, pp.
  7004--7014.

\bibitem[{Qiu et~al.(2021)Qiu, Zheng, Msahli, Memmi, Qiu, and
  Lu}]{Qiu2021TopologicalGC}
Qiu, H., Q.~Zheng, M.~Msahli, G.~Memmi, M.~Qiu, and J.~Lu, Topological Graph
  Convolutional Network-Based Urban Traffic Flow and Density Prediction.
  \emph{IEEE Transactions on Intelligent Transportation Systems}, Vol.~22,
  2021, pp. 4560--4569.

\bibitem[{Zhang et~al.(2022)Zhang, Sun, Guan, Chen, Witlox, and
  Huang}]{Zhang2022ForecastingTC}
Zhang, X., Y.~Sun, F.~Guan, K.~Chen, F.~Witlox, and H.~Huang, Forecasting the
  crowd: An effective and efficient neural network for citywide crowd
  information prediction at a fine spatio-temporal scale. \emph{Transportation
  Research Part C: Emerging Technologies}, 2022.

\bibitem[{Chen et~al.(2022{\natexlab{b}})Chen, Fu, and Wang}]{Chen2022AGC}
Chen, P., X.~Fu, and X.~Wang, A Graph Convolutional Stacked Bidirectional
  Unidirectional-LSTM Neural Network for Metro Ridership Prediction. \emph{IEEE
  Transactions on Intelligent Transportation Systems}, Vol.~23,
  2022{\natexlab{b}}, pp. 6950--6962.

\bibitem[{Huo et~al.(2023)Huo, Zhang, Wang, Gao, Hu, and
  Yin}]{Huo2023HierarchicalSG}
Huo, G., Y.~Zhang, B.~Wang, J.~Gao, Y.~Hu, and B.~Yin, Hierarchical
  Spatio–Temporal Graph Convolutional Networks and Transformer Network for
  Traffic Flow Forecasting. \emph{IEEE Transactions on Intelligent
  Transportation Systems}, Vol.~24, 2023, pp. 3855--3867.

\bibitem[{Yao et~al.(2023)Yao, Gu, Su, and Guizani}]{Yao2023MVSTGNAM}
Yao, Y., B.~Gu, Z.~Su, and M.~Guizani, MVSTGN: A Multi-View Spatial-Temporal
  Graph Network for Cellular Traffic Prediction. \emph{IEEE Transactions on
  Mobile Computing}, Vol.~22, 2023, pp. 2837--2849.

\bibitem[{Li et~al.(2023{\natexlab{b}})Li, Wang, Zhao, Yu, Hu, Yin, and
  Liu}]{Li2023IGNetAI}
Li, P., S.~Wang, H.~Zhao, J.~Yu, L.~Hu, H.~Yin, and Z.~Liu, IG-Net: An
  Interaction Graph Network Model for Metro Passenger Flow Forecasting.
  \emph{IEEE Transactions on Intelligent Transportation Systems}, Vol.~24,
  2023{\natexlab{b}}, pp. 4147--4157.

\bibitem[{Ji et~al.(2023)Ji, Wang, Huang, Wu, Xu, Wu, Zhang, and
  Zheng}]{ji2023spatio}
Ji, J., J.~Wang, C.~Huang, J.~Wu, B.~Xu, Z.~Wu, J.~Zhang, and Y.~Zheng,
  Spatio-Temporal Self-Supervised Learning for Traffic Flow Prediction. In
  \emph{Proceedings of the AAAI Conference on Artificial Intelligence}, 2023.

\bibitem[{Jiang et~al.(2023)Jiang, Wang, Yong, Jeph, Chen, Kobayashi, Song,
  Fukushima, and Suzumura}]{jiang_spatio-temporal_2023}
Jiang, R., Z.~Wang, J.~Yong, P.~Jeph, Q.~Chen, Y.~Kobayashi, X.~Song,
  S.~Fukushima, and T.~Suzumura, Spatio-Temporal Meta-Graph Learning for
  Traffic Forecasting. In \emph{Proceedings of the AAAI Conference on
  Artificial Intelligence}, 2023.

\bibitem[{Noursalehi et~al.(2018)Noursalehi, Koutsopoulos, and
  Zhao}]{Noursalehi2018RealTT}
Noursalehi, P., H.~N. Koutsopoulos, and J.~Zhao, Real time transit demand
  prediction capturing station interactions and impact of special events.
  \emph{Transportation Research Part C: Emerging Technologies}, 2018.

\bibitem[{Cheng et~al.(2021)Cheng, Tr{\'e}panier, and
  Sun}]{Cheng2021RealTimeFO}
Cheng, Z., M.~Tr{\'e}panier, and L.~Sun, Real-Time Forecasting of Metro
  Origin-Destination Matrices with High-Order Weighted Dynamic Mode
  Decomposition. \emph{Transp. Sci.}, Vol.~56, 2021, pp. 904--918.

\bibitem[{He et~al.(2015)He, Zhang, Ren, and Sun}]{He2015DeepRL}
He, K., X.~Zhang, S.~Ren, and J.~Sun, Deep Residual Learning for Image
  Recognition. \emph{2016 IEEE Conference on Computer Vision and Pattern
  Recognition (CVPR)}, 2015, pp. 770--778.

\bibitem[{Wang et~al.(2023)Wang, Jiang, Jiang, Han, and
  Zhao}]{Wang2023TowardsEA}
Wang, J., J.~Jiang, W.~Jiang, C.~Han, and W.~X. Zhao, Towards Efficient and
  Comprehensive Urban Spatial-Temporal Prediction: A Unified Library and
  Performance Benchmark. \emph{ArXiv}, Vol. abs/2304.14343, 2023.

\end{thebibliography}

\end{document}